# Guided Depth Map Super-Resolution via Multi-Scale Fusion U-shaped Mamba Network

Chenggang Guo[a,*], Hao Xu[a,**], XianMing Wan[b,***]

[a] *School of Computer and Software Engineering, Xihua University, Chengdu, China, 610039*
[b] *Shenzhen Adaps Photonics Technology Co., Ltd, Shenzhen, China, 518055*
\* e-mail: chenggang.guo90@hotmail.com (Corresponding Author)
\*\* e-mail: xuhao95@stu.xhu.edu.cn
\*\*\* e-mail: wanxm0307@163.com

**Abstract**—Depth map super-resolution technology aims to improve the spatial resolution of low-resolution depth maps and effectively restore high-frequency detail information. Traditional convolutional neural network has limitations in dealing with long-range dependencies and are unable to fully model the global contextual information in depth maps. Although transformer can model global dependencies, its computational complexity and memory consumption are quadratic, which significantly limits its ability to process high-resolution depth maps. In this paper, we propose a multi-scale fusion U-shaped Mamba (MSF-UM) model, a novel guided depth map super-resolution framework. The core innovation of this model is to integrate Mamba's efficient state-space modeling capabilities into a multi-scale U-shaped fusion structure guided by a color image. The structure combining the residual dense channel attention block and the Mamba state space module is designed, which combines the local feature extraction capability of the convolutional layer with the modeling advantage of the state space model for long-distance dependencies. At the same time, the model adopts a multi-scale cross-modal fusion strategy to make full use of the high-frequency texture information from the color image to guide the super-resolution process of the depth map. Compared with existing mainstream methods, the proposed MSF-UM significantly reduces the number of model parameters while achieving better reconstruction accuracy. Extensive experiments on multiple publicly available datasets validate the effectiveness of the model, especially showing excellent generalization ability in the task of large-scale depth map super-resolution.

## 1 INTRODUCTION

Depth map plays a key role in the field of 3D visual perception and are widely used in scenarios such as autonomous driving [1], mixed reality [2], and 3D reconstruction [3]. However, due to the cost of sensor hardware, physical limitations (such as the optical diffraction limit), and the complexity of the imaging environment, depth maps acquired by consumer-grade depth sensors (eg. structured light and time-of-flight methods) usually face problems such as low resolution, severe noise interference, and lack of detail information. These problems seriously restrict the accuracy and robustness of subsequent 3D vision tasks. Depth map super-resolution (DSR) technology aims to reconstruct low-resolution (LR) depth maps into high-resolution (HR) depth maps through algorithms. It is a key way to improve the quality of depth data and unleash its application potential.

In recent years, depth map super-resolution methods based on convolutional neural networks (CNNs) have gained widespread attention. However, the inherent local receptive field characteristics of CNNs make it difficult to effectively model the long-range dependencies and global context information that are widely present in images. This limitation may lead to blurred edges and structural distortions in the reconstructed depth map, especially when dealing with complex textures or large-scale super-resolution. Although the transformer model [4] can effectively capture global information through the self-attention mechanism, its quadratic computational complexity and large number of parameters make it face severe computational and memory overhead challenges when processing high-resolution images, limiting its application in practical deployment.

Recently, state space models (SSMs), especially the Mamba model [5], have made breakthrough progress in the field of long sequence modeling. By introducing input-dependent selective state transfer, Mamba achieves context modeling capabilities comparable to Transformer while maintaining linear computational complexity, providing a new and efficient solution for computer vision tasks. Its characteristics are particularly suitable for capturing the spatial global information of depth images. On the other hand, using high-resolution color (RGB) images to guide depth map super-resolution (GDSR) has become the mainstream paradigm. The rich texture and edge information in the color image provides a strong prior for detail recovery in the depth map. Effectively and fully integrating cross-modal information is the core challenge of GDSR methods.

This paper aims to explore a GDSR method that considers high performance, high model efficiency and strong cross-modal fusion capabilities. The core idea is to combine the Mamba module with the mature UNet architecture and multi-scale cross-modal fusion strategy. The main contributions are described as follows:

(1) A novel color map guided depth map super-resolution network named MSF-UM is proposed. Its core innovation lies in embedding the efficient Mamba state space module into the UNet architecture and constructing the RDCB-Mamba basic module to enhance the global context modeling capability based on local feature extraction.

(2) An efficient multi-scale cross-modal fusion mechanism is designed. MSF-UM adopts the UNet structure and uses the RDCB-Mamba module at the key nodes of the encoder and decoder to achieve multi-level and multi-scale deep fusion of depth map features and color map guided features.

(3) A better lightweight model is achieved. MSF-UM significantly reduces the number of model parameters and achieved a better balance between performance and efficiency. Experimental results show that our method achieves superior results in the field of guided depth map super-resolution compared to existing methods.

## 2 RELATED WORK

Using images to guide depth map reconstruction is the current mainstream research direction of depth map super-resolution. It is based on the assumption that there is a correlation between color image edges and depth edges, and the intensity changes of color edges often correspond to changes in depth discontinuous areas. Therefore, by obtaining color and depth images in the same field of view through camera calibration and registration technology, high-quality depth map reconstruction can be achieved by using the edge correlation of color and depth.

### 2.1 Traditional Methods

Filtering-based methods use filtering techniques to fuse the structural information of color and depth images, aiming to transfer color image details to depth images. Johannes et al. [6] used a joint bilateral filter to guide low-quality depth maps to high quality with the help of color images. In view of the inherent noise in the depth data obtained by TOF depth cameras or Kinect cameras, Wang et al. [7] optimized the low-resolution depth map by iterative joint bilateral filtering based on the color similarity between the center pixel and the neighboring pixels. Similarly, Chan et al. [8] designed a noise-aware filter, combined the scene depth value as a reference for geometric smooth areas, and applied an adaptive multi-edge upsampling filter to improve the quality of the depth map. He et al. [9] proposed a color image guided filter to smooth the edges of objects by guiding the image content and calculating the filter output. In order to reduce the halo artifacts in the scene depth map, Kou et al. [10] and Ochotorena et al. [11] introduced a gradient domain guided filter to better preserve the edges of the depth map by strengthening the first-order edge detection constraints. However, these filter-based depth map super-resolution methods tend to remove high-frequency information in the depth map when improving the resolution, resulting in blurred edge structures in the reconstructed depth map.

The core idea of the optimization-based method is to model the depth map super-resolution process as an optimization problem. This type of method regards the depth map super-resolution task as a convex optimization problem. It attempts to constrain the depth map upsampling process by obtaining the relevant structural information between the color image and the depth image with the help of carefully designed optimization functions and regularization terms. Diebel et al. [13] first applied the Markov random field to multimodal datasets to fill the gap between resolutions, but the reconstruction results obtained by this method are too smooth and lose a lot of detail information. Park et al. [14] combined multiple weighting factors and non-local mean filtering to keep the depth boundary clear in the depth map reconstruction results. Ferstl et al. [15] used higher-order regularization constraints to upsample depth images. Gu et al. [16] proposed a weighted analysis representation model and used a task-driven training strategy to learn the parameters in the task. Although these methods consider the similarity between depth maps and color images to a certain extent, due to the high algorithm complexity and the lack of universality of the manually designed optimization functions, different functions have different effects on the reconstruction results, which limits their further application.

### 2.2 Deep Learning based Methods

With the rapid development of deep learning technology, deep learning-based methods have made significant breakthroughs in the field of color image GDSR. The pixel adaptive convolution operation proposed by Su et al. [17] uses the idea of bilateral filter to expand the standard convolution operation, which can more effectively learn and utilize the guidance information of color images to help restore depth maps. Hui et al [18] constructed a multi-scale guided convolution network named MSGNet, which allows the LR depth map to be upsampled step by step under the guidance of the color image. He et al. [19] created a new dataset called "RGBDD (RGB-Depth-Depth)" and proposed an efficient and lightweight depth map super-resolution model. This method uses multiple Octave convolutions [20] to decompose high-resolution color features in stages to obtain high-frequency and low-frequency components, and then merges the high-frequency color features of each stage with the depth features of different levels to gradually enrich the details of the depth map features. Zhao et al. [21] constructed a discrete cosine network to extract shared and specific multimodal information through a semi-decoupled feature extraction module. Mallick et al. [22] innovatively integrated the advantages of CNN and Transformer encoders and used a local image attention mechanism to solve the GDSR problem. Ariav et al. [23] proposed a depth map super-resolution method based on a cascaded Transformer architecture. Its core innovation is to achieve multi-scale feature fusion of the HR color image and the LR depth map through cross-modal attention guidance. Yuan et al. [24] designed a recurrent structured attention guided network. This method gradually optimizes the edges and geometric structures of the depth map through a recurrent multi-stage architecture. Each stage contains a structural attention module, which uses the semantic information of the color image to generate a spatially adaptive weight map, dynamically fussing the high-frequency details and low-frequency contours of the depth features. The recurrent residual connection is introduced to strengthen the global structural consistency by reusing features across stages. Shi et al. [25] introduced a symmetric uncertainty method that selects appropriate color information to effectively restore HR depth map while skipping harmful textures. Sun et al. [26] developed cross-task knowledge distillation to exchange the correlation between the depth image super-resolution and depth estimation branches. However, existing deep learning based methods do not fully consider the multi-scale feature fusion structure that can be used between the color image and the depth map. By introducing the popular UNet encoder-decoder structure, this paper designs a multi-scale fusion architecture and uses a novel Mamba based module to realize the multimodal feature fusion and information interaction.

## 3 METHOD

In this section, we present the details of the proposed Multi-Scale Fusion U-shaped Mamba (MSF-UM) model. We first describe the general optimization problem of the GDSR task. Then, we introduce the network structure of the MSF-UM model and the proposed key components.

### 3.1 Probelm Formulation

Given an input LR depth map $D_{lr} \in R^{h \times w \times 1}$ and a HR RGB color image $I_{rgb} \in R^{sh \times sw \times 3}$, GDSR aims to predict a HR depth map under the supervision of a HR groundtruth depth map $D_{hr} \in R^{sh \times sw \times 1}$, where $h$, $w$, $s$ denotes the height, width, and scaling factor, respectively. In general, a GDSR network is input with a paired $D_{lr}$ and $I_{rgb}$, and is trained to generate a HR output $D_{sr}$ close to a given target $D_{hr}$. The process can be formulated as follows:

$$D_{hr} \approx D_{sr} = f(D_{lr}, I_{rgb}; \theta) \qquad (1)$$

where $f$ is the GDSR model that learns the nonlinear mapping from $D_{lr}$ to $D_{hr}$, and $\theta$ represents the learning parameters of $f$.

### 3.2 Network Structure

MSF-UM is a depth map super-resolution deep model that integrates UNet architecture, color image multi-scale guidance, Mamba-based residual dense channel-attention module. It aims to improve the reconstruction quality of depth images through multi-level feature extraction, global context modeling and multi-scale fusion of cross-modal information. The core design idea of this model is to embed the Mamba module and RDCB module into the encoder-decoder framework of UNet, and combine the color image multi-scale guidance to make full use of the multi-scale feature extraction ability of UNet, the long-range dependency modeling advantage of Mamba and the high-frequency texture information in the color image. Through this collaborative design, MSF-UM can not only effectively capture the local details and global structure of depth images, but also further enhance the reconstruction effect through the guidance information of color images, especially in complex scenes and large-scale super-resolution tasks. Figure 1 shows the overall network structure of MSF-UM. The network structure design of MSF-UM and the functions and advantages of its key components will be introduced in detail below.

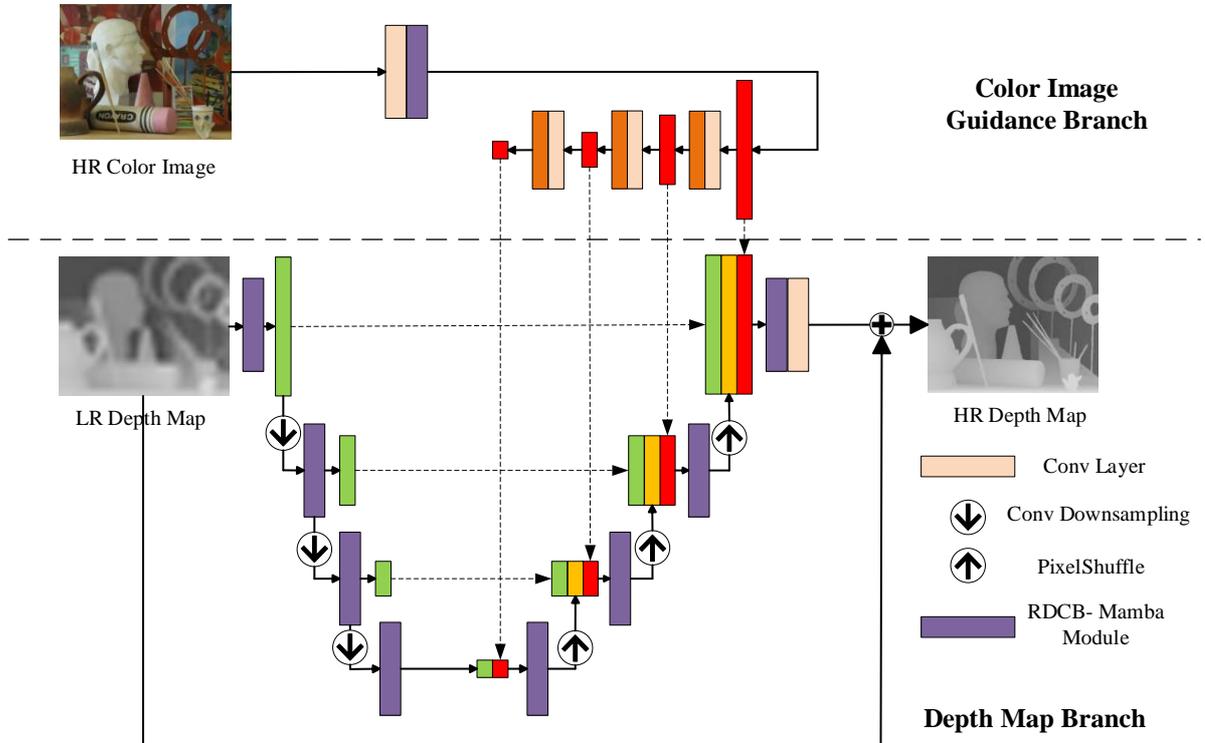

Figure 1 Network structure of the proposed Multi-Scale Fusion U-shaped Mamba model (MSF-UM).

The overall design of MSF-UM adopts a cross-modal guided fusion strategy of a U-shaped multi-scale structure. Its depth map branch is based on the classic encoder-decoder framework of UNet, which consists of symmetrical downsampling paths and upsampling paths, forming a U-shaped network structure.

In the encoder part, i.e. the depth map multi-scale downsampling module, the model gradually extracts the features of the input depth map by introducing a novel combination of residual dense channel attention block (RDCB) and Mamba module. The RDCB module enhances the representation ability of features through dense connections and channel attention mechanism, while the Mamba module captures long-range dependencies in the depth image through the selective state space mechanism, further improving the global context modeling ability. During the downsampling process, the model gradually reduces the size of the feature map through convolution downsampling operations with a stride of 2, while increasing the number of channels to capture features of different scales.

In the decoder part, i.e., the feature multi-scale fusion upsampling module, the model gradually restores the size of the feature map through upsampling operations, and combines the RDCB-Mamba module at each scale to perform cross-modal feature fusion of features from encoder skip connections, upsampled features, and color map guided features. This design enables the model to make full use of the multi-level feature information extracted from the encoder and the color image guidance branch to gradually restore the details and structures of high-resolution images. The skip connection mechanism plays an important role in the decoder, which directly transfers the shallow features of the corresponding scale in the encoder to the decoder, avoiding information loss. Through this multi-level feature extraction and fusion strategy, MSF-UM can restore accurate details while maintaining the consistency of the overall structure of the depth image, especially in the edge and texture areas.

The color image guidance branch introduces the RDCB-Mamba module to extract features from the input color image to fully utilize the high-frequency texture information in the color image. Before the downsampling step, the RDCB-Mamba module is used to process the high-resolution color image information to enhance the modeling capability of global context information. This module dynamically filters irrelevant information through the selective scan algorithm, retains key features, and uses the extracted features for fusion with the depth image features. This design enables the model to better utilize the texture information of the color image and further enhance the reconstruction quality of the depth image.

### 3.3 RDCB-Mamba module

As shown in Figure 2, Residual Dense Channel-attention Block (RDCB) from MSF-UNet [12] achieves efficient feature extraction and fusion through the combination of dense connection, residual structure, and channel attention mechanism. RDCB contains six densely connected convolutional layers, which integrate multi-level semantic features through cascade operation, making full use of the complementarity of features at different semantic levels. After feature integration, the module fuses the cascaded features through a convolutional layer to reduce the feature dimension and enhance the global consistency of the features. Subsequently, the channel attention mechanism is introduced to dynamically screen the fused features. By calculating the importance weights of each channel, the high-frequency information that is critical to the reconstruction task is highlighted, while redundant or irrelevant features are suppressed, thereby improving the representation ability of the features. Finally, the module adds the initial input features to the processed features through residual connection. This design not only ensures the stability of the basic structure, but also enhances the recovery ability of high-frequency details through residual learning. Through this multi-level feature reuse, RDCB provides strong support for the reconstruction of high-quality depth maps.

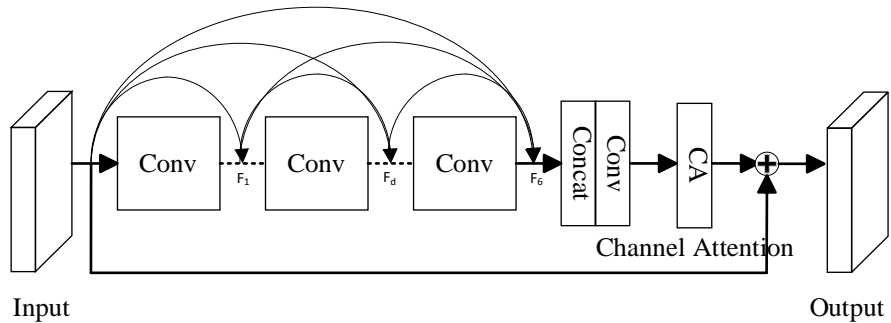

Figure 2 Residual Dense Channel-attention Block

Mamba reduces the computational complexity from the quadratic level of the classic self-attention mechanism to the linear level through a selective state space mechanism. This design enables Mamba to significantly reduce computational overhead and memory usage when processing high-resolution images while maintaining the ability to efficiently model global context information. Assuming that the size of the input feature map is $H \times W \times C$ (height, width, number of channels), Mamba first flattens it into a sequence of length $L=H \times W$, and the dimension of each sequence element is $C$. This serialization operation enables Mamba to convert the spatial information of the image into sequence data, which is convenient for subsequent state space modeling. After layer normalization, these features enter the Mamba block containing two parallel branches. In the first branch, the sequence features are expanded to $2L$ through a linear layer, followed by a one-dimensional convolutional layer, SiLU activation function, and SSM layer. In the second branch, these sequence features are also expanded to $2L$ by a linear layer, followed by SiLU activation function. After that, the features from the two branches are multiplied and merged. Finally, the features are projected to the original length $L$, and then reshaped into a feature map of $H \times W \times C$ for subsequent network layers.

The RDCB-Mamba module consists of two RDCB modules and two Mamba modules stacked together. Figure 3 shows the module structure. Specifically, the module first extracts local features through two consecutive RDCB modules and uses the channel attention mechanism to filter and enhance the features. Subsequently, the Mamba module is used to perform global context modeling on the processed features to capture long-range dependencies in the image. Through this design, the RDCB-Mamba module can not only effectively capture the local details and global structure of the image, but also ensure that the model can fully utilize the global information of the image through multi-level feature reuse and dynamic feature screening, thereby improving the model's expressiveness.

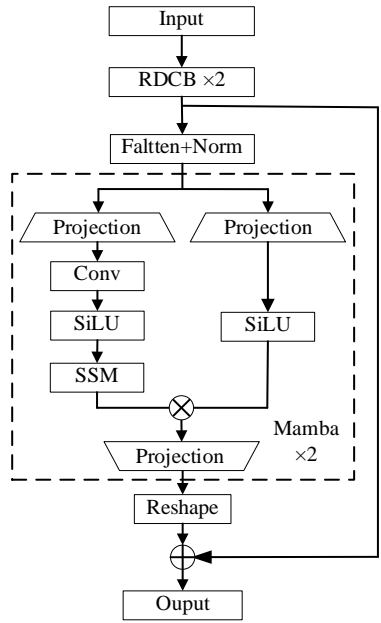

Figure 3 RDCB-Mamba Block

## 4 EXPERIMENT
In this section, comprehensive quantitative and qualitative experiments are conducted on multiple datasets to demonstrate the effectiveness of the proposed MSF-UM model.

### 4.1 Implementation Details
In order to comprehensively evaluate the performance of the proposed method, this section conducts sufficient experimental verification on four widely used public datasets, including NYUv2 [27], Middlebury [28], Lu [29], and RGBDD [19]. Specifically, the NYUv2 dataset is selected as the main training and validation dataset. The dataset contains 1449 pairs of RGB-D images, of which the first 1000 pairs are used for model training and the remaining 449 pairs are used for verification. In order to further verify the generalization ability of the model, we also tests on the Middlebury dataset (30 pairs), Lu dataset (6 pairs), and RGBDD dataset (405 pairs). These three datasets cover a variety of indoor and outdoor scenes to further expand the coverage of the experiments. During the training process, high-resolution (HR) depth images are cropped into patches of size 256×256 to ensure that the model can fully learn local detail information and speed up training. The low-resolution (LR) depth images used in the experiment are generated by bilateral downsampling to simulate the degradation process in real scenes. This multi-dataset, multi-scenario experimental design not only verifies the performance of the model on known data, but also fully tests its generalization ability in different environments.

The root mean square error (RMSE) is used as an evaluation metric to quantify the difference between the reconstructed depth map and the groundtruth map. The lower the RMSE value, the higher the reconstruction quality and the closer the geometric structure and detail restoration of the depth map are to the real scene. To ensure the repeatability and efficiency of the experiment, the model is implemented based on the Pytorch framework [30] and trained on the NVIDIA A40 GPU. During the training process, the batch size is set to 2 to balance memory usage and training stability. The optimizer is Adam [31], and its hyperparameters are set to $\beta_1=0.9$, $\beta_2=0.999$, and $\varepsilon_1=0.9$. The initial learning rate is set to $1e^{-4}$ and is decayed by 0.1 after every 150 epochs.

### 4.2 Experimental Results
In order to fully demonstrate the superiority of the MSF-UM model, this study compares it with several current leading methods in both quantitative and qualitative aspects. The quantitative comparison is mainly based on indicators such as the root mean square error (RMSE) to objectively evaluate the difference between the reconstructed depth map and the groundtruth map. The qualitative comparison shows the performance of the algorithm in detail recovery, edge clarity, and texture consistency through visualization results.

### 4.2.1 Quantitative comparison
Quantitative experiments are conducted on four widely used datasets NYUv2, Middlebury, Lu, and RGBDD, and tested with super-resolution scale factors of 4, 8, and 16, respectively. The comparison results are shown in Tables 1 and 2, with the best performance shown in bold font and the second best shown in underline.

From the quantitative results in Tables 1 and 2, it can be seen that MSF-UM performs well on multiple datasets and different upsampling scales, achieving the best or second-best performance, and even surpasses existing methods at large scaling factors. Especially in large-scale super-resolution tasks with a scaling factor of 16, the proposed method achieved the best results on all datasets, which fully verified the effectiveness of the proposed RDCB-Mamba module and multi-scale feature fusion strategy. It is particularly noteworthy that some existing methods only perform well on specific datasets or specific scale factors, while MSF-UM shows stronger adaptability and stability, and can maintain excellent reconstruction effects in a variety of scenarios and tasks. This performance consistency across datasets and scale factors not only verifies the advantages of MSF-UM in GDSR tasks, but also fully demonstrates its strong generalization ability.

Table 1 Quantitative comparison on NYUv2 and Middlebury datasets (average RMSE).

| Methods | NYUv2 | | | Middlebury | | |
|---|---|---|---|---|---|---|
| | ×4 | ×8 | ×16 | ×4 | ×8 | ×16 |
| Bicubic | 8.16 | 14.12 | 22.32 | 2.28 | 3.98 | 6.37 |
| DJF[32] | 2.80 | 5.33 | 9.46 | 1.68 | 3.24 | 5.62 |
| DJFR[33] | 2.38 | 4.94 | 9.18 | 1.32 | 3.19 | 5.57 |
| DKN[34] | 1.62 | 3.26 | 6.51 | 1.23 | 2.12 | 4.24 |
| FDKN[34] | 1.86 | 3.58 | 6.96 | 1.08 | 2.17 | 4.50 |
| CUNet[35] | 1.92 | 3.70 | 6.78 | 1.10 | 2.17 | 4.33 |
| FDSR[19] | 1.61 | 3.18 | 5.86 | 1.13 | 2.08 | 4.39 |
| DCTNet[21] | 1.59 | 3.16 | 5.84 | 1.10 | 2.05 | 4.19 |
| SUFT[25] | **1.12** | 2.51 | 4.86 | **1.07** | **1.75** | 3.18 |
| MSF-UM | 1.15 | **2.48** | **4.67** | 1.19 | 1.82 | **3.14** |

Table 2 Quantitative comparison on Lu and RGBDD datasets (average RMSE).

| Methods | Lu | | | RGBDD | | |
|---|---|---|---|---|---|---|
| | ×4 | ×8 | ×16 | ×4 | ×8 | ×16 |
| DJF[32] | 1.65 | 3.96 | 6.75 | 3.41 | 5.57 | 8.15 |
| DJFR[33] | 1.15 | 3.57 | 6.77 | 3.35 | 5.57 | 7.99 |
| DKN[34] | 0.96 | 2.16 | 5.11 | 1.30 | 1.96 | 3.42 |
| FDKN[34] | **0.82** | 2.10 | 5.05 | 1.18 | 1.91 | 3.41 |
| CUNet[35] | 0.91 | 2.23 | 4.99 | 1.18 | 1.95 | 3.45 |
| FDSR[19] | 1.29 | 2.19 | 5.00 | 1.16 | 1.82 | 3.06 |
| DCTNet[21] | 0.88 | 1.85 | 4.39 | **1.08** | 1.74 | 3.05 |
| SUFT[25] | 1.10 | **1.74** | 3.92 | 1.10 | **1.69** | 2.71 |
| MSF-UM | 1.05 | 1.82 | **3.82** | 1.17 | 1.70 | **2.68** |

### 4.2.2 Qualitative comparison on the validation set

In order to demonstrate the performance of MSF-UM, a qualitative comparison is first conducted on the NYUv2 dataset. The experimental results are shown in Figure 4. From the perspective of qualitative analysis, MSF-UM shows significant advantages in depth super-resolution. Its reconstructed depth maps not only have smaller prediction deviations, but also presents finer edges and details in terms of visual effects. Compared with existing methods, the depth maps generated by MSF-UM has fewer artifacts. In addition, compared with the ground truth (GT) depth maps, the reconstruction results of MSF-UM show high consistency, especially in areas with complex textures and interlaced edges. It can better preserve the geometric structure and detail of the scene. These results show that MSF-UM has advantages in processing large-scale depth map super-resolution tasks,.

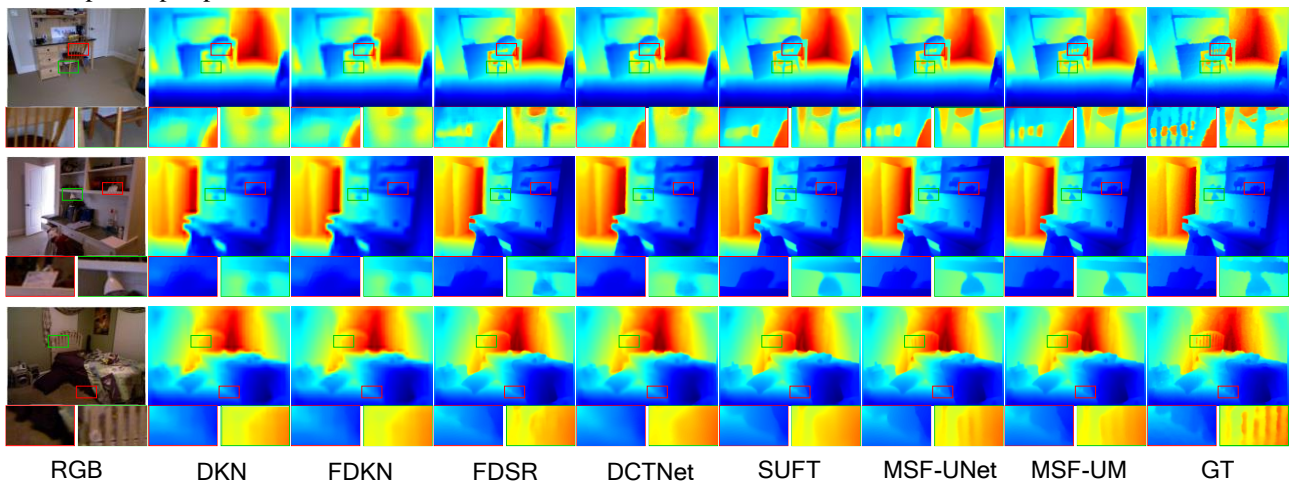

Figure 4 Visual comparison of ×16 super-resolution results on the NYUv2 dataset.

### 4.2.3 Generalization comparison on the test set

In order to fully verify the generalization ability of the MSF-UM model, generalization comparison are conducted on three widely used test sets: Middlebury, Lu, and RGBDD. These datasets cover a variety of scenes and complex texture information, which can effectively test the adaptability and robustness of the model in different environments. The visualization comparison of ×16 super-resolution results is shown in Figure 5.

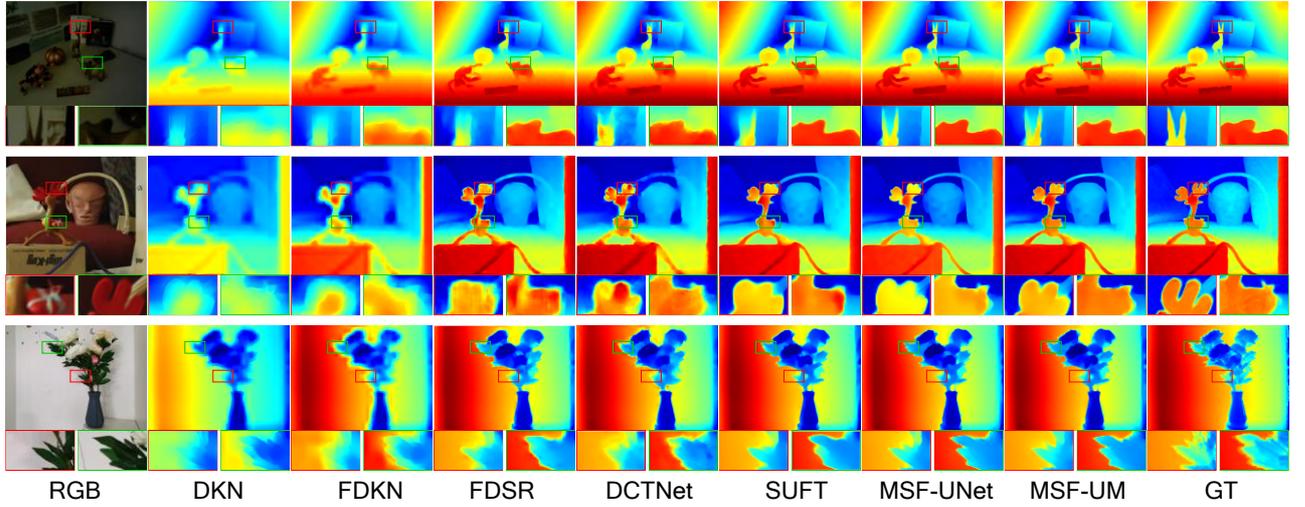

Figure 5 Visual comparison of ×16 super-resolution results on three test datasets.

Experimental results show that MSF-UM demonstrates excellent reconstruction performance on multiple unseen test sets, and its generated high-resolution depth maps outperform existing methods in terms of visual quality. Compared with existing methods, MSF-UM can effectively reduce the generation of artifacts while minimizing the introduction of noise, making the reconstruction results more natural and realistic. MSF-UM can accurately capture and restore the geometric structure and detail information of the scene, especially in high-frequency details and complex texture areas.

### 4.3 Parameter Comparison

To further evaluate the balance between model size and performance, Table 3 shows the comparison of the number of learnable parameters and RMSE of each model under three scale factors of ×4, ×8 and ×16 on the NYUv2 dataset. It shows that MSF-UM achieves better or comparable performance with relatively fewer parameters under multiple scale factors than existing models. The efficient design proposed in this paper not only makes the model more practical in environments with limited computing resources, but also provides new ideas for developing lightweight networks, especially in scenarios that require real-time processing and high-precision reconstruction, such as autonomous driving, virtual reality, and 3D reconstruction.

Table 3 The number of parameters and RMSE of ×4, ×8, and ×16 GDSR models on the NYUv2 dataset

| Methods | NYUv2 ×4 | | NYUv2 ×8 | | NYUv2 ×16 | |
|---|---|---|---|---|---|---|
| | RMSE | Params(M) | RMSE | Params(M) | RMSE | Params(M) |
| DJF[32] | 2.80 | 0.05 | 5.33 | 0.05 | 9.46 | 0.05 |
| DJFR[33] | 2.38 | 0.055 | 4.94 | 0.055 | 9.18 | 0.055 |
| DKN[34] | 1.62 | 1.16 | 3.26 | 1.16 | 6.51 | 1.16 |
| FDKN[34] | 1.86 | 0.69 | 3.58 | 0.69 | 6.96 | 0.69 |
| CUNet[35] | 1.92 | 0.2 | 3.70 | 0.2 | 6.78 | 0.2 |
| FDSR[19] | 1.61 | 0.6 | 3.18 | 0.6 | 5.86 | 0.6 |
| DCTNe[21] | 1.59 | 0.48 | 3.16 | 0.48 | 5.84 | 0.48 |
| SUFT[25] | 1.12 | 22.01 | 2.51 | 39.95 | 4.86 | 97.36 |
| MSF-UM | 1.16 | 1.27 | 2.48 | 1.74 | 4.67 | 2.12 |

### 4.4 Ablation Study

In order to verify the effectiveness of each key module in the MSF-UM network, an ablation experiment was carried out on the NYUv2 and RGBDD datasets for the ×16 super-resolution task. The experimental results are shown in Table 4. First, a baseline model is constructed as a comparison by removing the RDCB module, the Mamba module, and the color image multi-scale guidance branch. It shows that the super-resolution RMSE value of the baseline model is high and the reconstruction effect is limited. When the color map multi-scale guidance branch, RDCB module, and Mamba module are gradually added to the baseline network, a significant improvement in model performance can be observed. The synergy of these modules enables MSF-UM to perform well in the GDSR task, verifying the rationality and effectiveness of the module design proposed in this paper.

Table 4 Ablation study of MSF-UM on NYUv2 and RGBDD datasets (average RMSE, ×16)

| Baseline model | Color image guidance branch | RDCB | Mamba | NYUv2 | RGBDD |
|---|---|---|---|---|---|
| ✓ | | | | 7.29 | 3.57 |
| ✓ | ✓ | | | 5.85 | 3.04 |
| ✓ | ✓ | ✓ | | 5.27 | 2.91 |
| ✓ | ✓ | | ✓ | 5.15 | 2.82 |
| ✓ | ✓ | ✓ | ✓ | **4.67** | **2.68** |

## CONCLUSIONS

In this paper, we propose a novel and efficient color map guided depth map super-resolution network, MSF-UM. By strategically integrating a selective state-space model into a color map guided multi-scale UNet architecture, MSF-UM effectively addresses the limitations of CNN in capturing long-range dependencies while avoiding the computational burden of the Transformer. Its core innovation lies in the RDCB-Mamba module, which enhances local feature representations through a residual dense channel attention mechanism and leverages Mamba for efficient global context modeling, thereby achieving a synergistic effect. Experiments on multiple benchmark datasets demonstrate that MSF-UM achieves reconstruction quality close to or reaching the state-of-the-art level at different super-resolution scales with very few model parameters. Both quantitative metrics (RMSE) and qualitative visualization results confirm that MSF-UM is capable of reconstructing depth maps with fine details, and geometrically consistent depth maps, especially in large-scale super-resolution (×16) tasks. MSF-UM has a potential for applications that require high-quality depth maps in resource-constrained environments (e.g. mobile AR/VR, robotic perception, and embedded 3D vision systems). In addition, Mamba has been successfully integrated into the mature UNet fusion framework, which also verified the effectiveness of this method in color image guided depth map super-resolution tasks. Future work can explore the optimization methods of vision Mamba in multimodal super-resolution tasks.

## FUNDING
No funding was received for conducting this study.

## CONFLICT OF INTERESTS
The authors declare they have no financial interests.

## COMPLIANCE WITH ETHICAL STANDARDS
This article is a completely original work of its authors; it has not been published before and will not be sent to other publications until the PRIA editorial board decides not to accept it for publication.